\documentclass[10pt,twocolumn,letterpaper]{article}

\usepackage{iccv}
\usepackage{times}
\usepackage{epsfig}
\usepackage{graphicx}
\usepackage{amsmath}
\usepackage{amssymb}

\newcommand{\figref}[1]{Fig.~\ref{#1}}
\newcommand{\tabref}[1]{Tab.~\ref{#1}}
\newcommand{\secref}[1]{Sec.~\ref{#1}}

\newcommand{\equref}[1]{Eq.~(\ref{#1})}

\newcommand{\tocite}[1]{\textcolor{red}{[TOCITE]}}

\newcommand{\important}[1]{{\textcolor{red}{[IMPORTANT!!!]}}}

\newcommand{\R}[1]{{%
    \textbf{%
        \ifstrequal{#1}{1}{\textcolor{Bittersweet}{R#1}}{%
        \ifstrequal{#1}{2}{\textcolor{ForestGreen}{R#1}}{%
        \ifstrequal{#1}{3}{\textcolor{cyan}{R#1}}{%
        \ifstrequal{#1}{4}{\textcolor{teal}{R#1}}{%
                           \textcolor{cyan}{R#1}%
        }}}}%
    }%
}}

\usepackage{hyperref}
\hypersetup{pagebackref=true,breaklinks=true,letterpaper=true,colorlinks,bookmarks=false}

\usepackage[accsupp]{axessibility}  

\iccvfinalcopy 

\def\iccvPaperID{5114} 

\ificcvfinal\fi

\begin{document}

\title{Hierarchical Generation of Human-Object Interactions \\ with Diffusion Probabilistic Models}

\author{Huaijin Pi$^{1}$,
~Sida Peng$^{1}$,
~Minghui Yang$^{2}$,
~Xiaowei Zhou$^{1}$,
~Hujun Bao$^{1}$\thanks{Corresponding author.}\\[1.5mm]
$^{1}$ State Key Lab of CAD\&CG, Zhejiang University \quad $^{2}$ Ant Group}

\maketitle
\ificcvfinal\thispagestyle{empty}\fi
\begin{abstract}
This paper presents a novel approach to generating the 3D motion of a human interacting with a target object, with a focus on solving the challenge of synthesizing long-range and diverse motions, which could not be fulfilled by existing auto-regressive models or path planning-based methods.
We propose a hierarchical generation framework to solve this challenge.
Specifically, our framework first generates a set of milestones and then synthesizes the motion along them. 
Therefore, the long-range motion generation could be reduced to synthesizing several short motion sequences guided by milestones.
The experiments on the NSM, COUCH, and SAMP datasets show that our approach outperforms previous methods by a large margin in both quality and diversity.
The source code is available on our project page \href{https://zju3dv.github.io/hghoi/index.html}{https://zju3dv.github.io/hghoi}.
\end{abstract}
\section{Introduction}
\label{sec:intro}
Scene-aware motion generation \cite{21iccv_samp} aims to synthesize 3D human motion given a 3D scene model to enable virtual humans to naturally wander around scenes and interact with objects, which has a variety of applications in AR/VR, filmmaking, and video games.

Unlike traditional motion generation methods for character control which aim to generate short or repeated motion on the fly guided by a user's control signals \cite{19tog_nsm},
we focus on the setting of generating long-term human-object interactions \cite{19tog_nsm, 21iccv_samp, 22eccv_couch} given a starting position of the human and a target object model. This setting brings in new challenges.
First, the entire approaching process and the human-object interaction should be coherent, which requires the capability of modeling long-range interaction between the human and the object.
%
Second, in the context of content generation, the generative model should be able to synthesize diverse motions as there are many plausible ways for a real human to approach and interact with the target object.
%
\begin{figure}[t]
  \centering
   \includegraphics[width=1.0\linewidth]{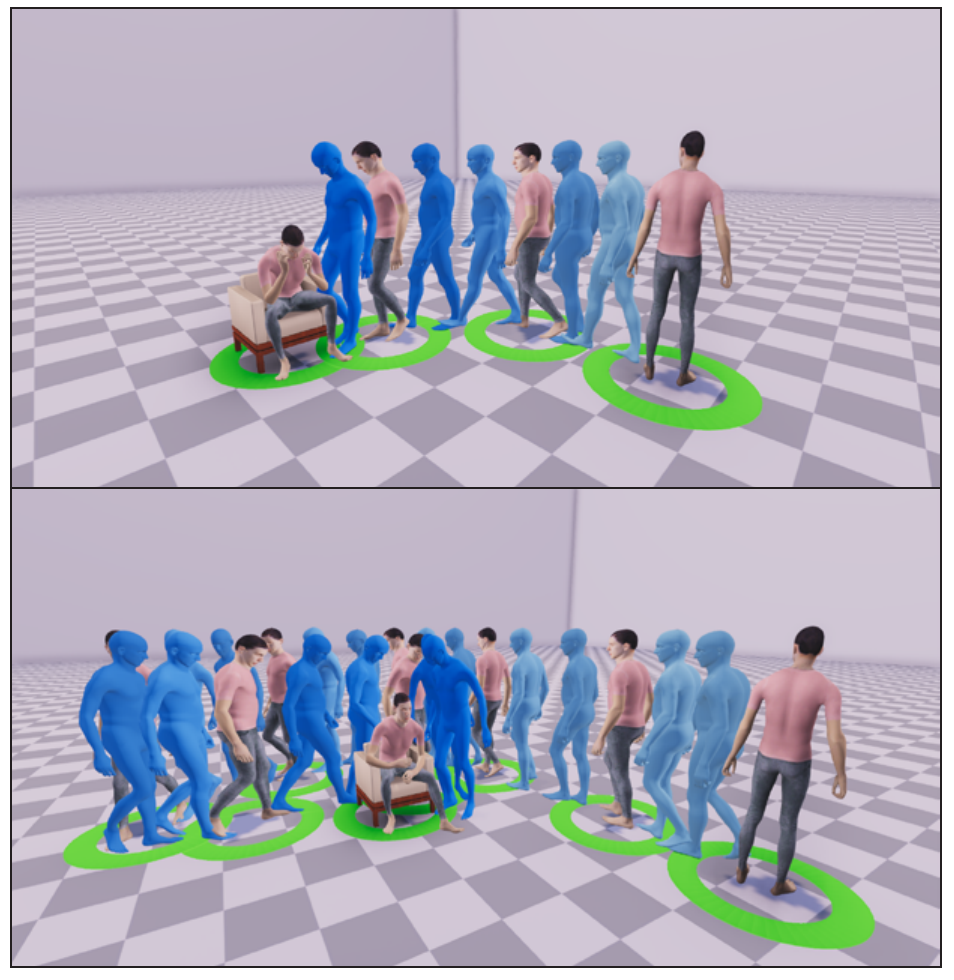}
   \caption{\textbf{Generation of human-object interactions.} Given an object, our method first predicts a set of milestones, where the rings indicate the positions and the humans with pink clothes represent the local poses. Then the motions are infilled between milestones. This figure shows that our method can generate diverse milestones and motions with the same object. The flow of time is shown with a color code where darker blue denotes the later frame.}
   \label{fig:basic}
\end{figure}

Existing methods for motion synthesis can be roughly characterized into online generation and offline generation.
Most online methods \cite{19tog_nsm, 21iccv_samp, 22eccv_couch} focus on real-time control of characters.
Given a target object, they generally use auto-regressive models to recurrently generate future motions by feeding their predictions.
Although they have been widely used for interactive scenarios like video games, their motion quality is not satisfactory enough for long-term generation \cite{22cvpr_tdns}.
A plausible cause is the error accumulation in the auto-regressive process, where the errors in previous predictions are fed back as the model input, as discussed in \cite{17cvpr_resrnn, 17tog_pfnn, 21iccv_actor,22eccv_temos, 22nips_nmf}.
To improve the motion quality, some recent offline methods \cite{20eccv_lhmp, 21cvpr_sgnhms, 21cvpr_slt, 22cvpr_tdns} employ a multi-stage framework, which first generates the trajectory and then synthesizes motions.
TDNS \cite{22cvpr_tdns} generates paths by combining the cVAE model \cite{14iclr_vae} and deterministic planning methods like A* \cite{68_astar}.
Although this strategy can produce reasonable paths, the path diversity is limited, as demonstrated by our experimental results in \secref{exp:ablation}.

In this paper, we propose a novel offline approach for synthesizing long-term and diverse human-object interactions.
Our innovation lies in a hierarchical generation strategy, which first predicts a set of milestones and then generates human motions between milestones.
\figref{fig:basic} illustrates the basic idea.
Specifically, given the starting position and the target object, we design a \textit{milestone generation module} to synthesize a set of milestones along the motion trajectory, each of which encodes the local pose and indicates the transition point during the human movement.
Based on these milestones, we employ a \textit{motion generation module} to produce the full motion sequence.
Thanks to the milestones, we simplify the long-sequence generation into synthesizing several short motion sequences.
Furthermore, the local pose at each milestone is generated by a transformer that considers the global dependency \cite{17nips_attention}, leading to temporally consistent results, which further contribute to coherent motions.

In addition to our hierarchical generation framework, we further exploit diffusion models \cite{15icml_diffusion, 20nips_ddpm, 20nips_song_score} to synthesize human-object interactions.
Previous diffusion models for motion synthesis \cite{22arxiv_flame, 22arxiv_hmd} combine transformer \cite{17nips_attention} and Denoising Diffusion Probabilistic Model (DDPM) \cite{20nips_ddpm}.
Directly applying them to our setting is prohibitively compute-intensive due to the long motion sequences and may lead to the GPU memory explosion \cite{22cvpr_stablediffusion}.
Because our hierarchical generation framework converts the long-term generation to the synthesis of several short sequences, the required GPU memory is reduced to the same level of short-term motion generation.
Therefore, we can efficiently leverage the transformer DDPM to synthesize long-term motion sequences, which improves the generation quality.

We validate our design choices on the NSM \cite{19tog_nsm}, COUCH \cite{22eccv_couch}, and SAMP \cite{21iccv_samp} datasets with extensive experiments. 
On these datasets, our hierarchical framework outperforms previous methods significantly in both motion quality and diversity.
\section{Related Work}
\label{sec:related}
\subsection{Motion synthesis}
\label{related:motion}
Motion synthesis is a long-standing problem in computer graphics and vision \cite{97cga_interpolation, 98cga_motionverb, 05tog_geostaticmotion, 16gdc_motionmatching}.
With the rapid development of deep learning, recent works have applied neural networks to motion synthesis \cite{15iccv_erd, 17cvpr_resrnn, 17tog_pfnn, 18tog_mann}. 
Some methods are deterministic \cite{16cvpr_strcnn, 18iclr_aclstm, 18bmvc_quaternet, 19iccv_ltd} while others try to predict stochastic motions by VAE \cite{18eccv_mtvae, 20cvpr_stochasticschememotion, 20eccv_dlow, 22cvpr_gamma, 21iccv_actor} and GAN \cite{18cvprw_hpgan, 19aaai_bihmpgan}.
To further improve the performance, some works apply GCN \cite{19iccv_ltd,20cvpr_ldr, 20eccv_his, 20cvpr_dmgnn, 21ijcv_multi} or transformers \cite{20eccv_lpj} to extract the features from the human skeleton.
To handle the ambiguity of human motion, some works \cite{17tog_pfnn, 22tog_deepphase} propose to employ phase signals to guide the motions. 
Recent works start to consider the scene context \cite{19tog_nsm, 20eccv_lhmp}.
NSM \cite{19tog_nsm} is the first work that aims at synthesizing human motions with object-level interactions with specific action labels.
Based on NSM, SAMP \cite{21iccv_samp} applies cVAE to predict diverse motions.
These works \cite{19tog_nsm, 21iccv_samp, 22eccv_couch} mainly focus on the interaction with one or two objects while others \cite{20eccv_lhmp, 21cvpr_sgnhms} generate motions with a full scene (including the information about floors and walls) as the input.
\cite{20eccv_lhmp} predicts future motions of the given motion sequence by first predicting the trajectory and then generating motions based on a 2D image of the scene.
Similar to this pipeline, \cite{21cvpr_sgnhms} applies GAN \cite{20commacm_gan}.
Furthermore, \cite{21cvpr_slt} proposes a framework that first places pose on a human-provided path and then synthesizes motions in a 3D scene with a full-scene scan.
\cite{22cvpr_tdns} combines $\text{A}^*$ and cVAE to generate diverse human motions.
To further control human motion, \cite{22eccv_gimo} employs the gaze, and COUCH \cite{22eccv_couch} explicitly models the hand contacts to guide the prediction.
Some methods \cite{21aaai_learn2sit, 23sig_physicalcsi} also enable physically simulated characters to perform scene interaction tasks including sitting \cite{21aaai_learn2sit, 23sig_physicalcsi} and carrying boxes \cite{23sig_physicalcsi}.
Some works also focus on grasp \cite{22cvpr_goal, 22eccv_saga}, manipulation \cite{21tog_manipnet} and the interaction with dynamic objects \cite{20tog_localphase}.
\paragraph{Ours vs. others.}
This work follows the setting of \cite{19tog_nsm, 21iccv_samp} and focuses on object-level interaction.
In contrast to \cite{19tog_nsm, 21iccv_samp} which generate motions mainly based on auto-regressive models, we design a hierarchical framework to synthesize motions.
Different from \cite{20eccv_lhmp, 21cvpr_slt}, our work focuses on much more long-term generation (longer than $10$ seconds) while \cite{20eccv_lhmp} is $2$ seconds and \cite{21cvpr_slt} is $6$ seconds.
Instead of planning the path auto-regressively with an extra network to generate diverse trajectories like \cite{22cvpr_tdns}, our method directly predicts a set of milestones to describe the approaching process which is inherently diverse.
In addition, most methods \cite{21iccv_samp, 22eccv_couch, 21cvpr_slt, 22cvpr_tdns} rely on cVAE to generate stochastic motions while we exploit DDPM \cite{20nips_ddpm} to synthesize trajectories and motions.

\begin{figure*}[t]
  \centering
   \includegraphics[width=1.0\linewidth]{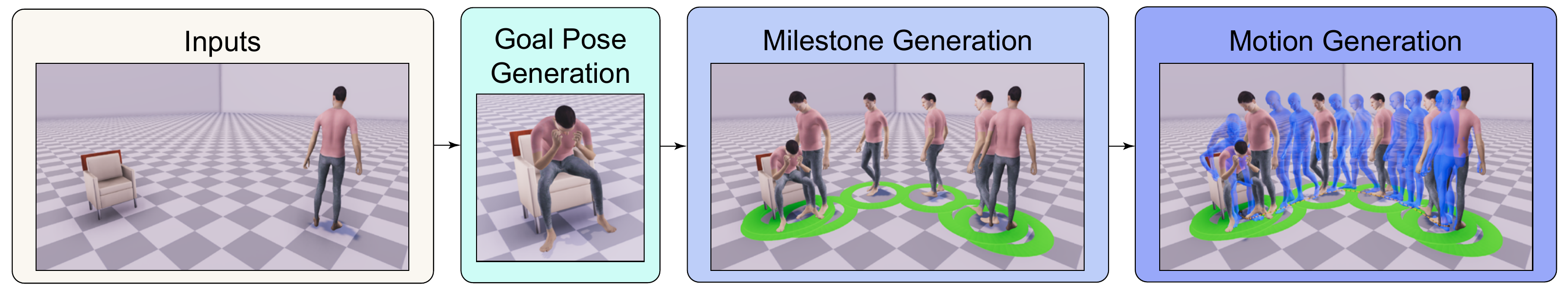}
   \caption{\textbf{Overview of our pipeline.} Our pipeline consists of three components: First, the goal pose is synthesized given the object. Then, a number of milestones with local poses are predicted based on the goal pose. Finally, the trajectory and the full motion sequences are infilled between the milestones.}
   \label{fig:overview}
\end{figure*}
\subsection{Diffusion models}
\label{related:diffusion}
Diffusion models \cite{15icml_diffusion} are a class of likelihood-based methods that generate samples by gradually removing noise from a signal.
Then, \cite{20nips_ddpm, 20nips_song_score} develop the diffusion models for high-quality image generation.
To control the generated results, \cite{21nips_diffusionbeatsgan} proposes classifier guidance for trading off diversity for fidelity. 
Later, the classifier-free model \cite{2022arxiv_classifierfree} achieves better results \cite{22arxiv_dalle2} in text-conditioned image generation.
In addition, diffusion models have been successfully applied to other domains like the generation of videos \cite{22arxiv_makeavideo, 22arxiv_videodiffusion} and 3D contents \cite{22arxiv_dreamfusion}.

There are some works \cite{22arxiv_motiondiffuse, 22arxiv_hmd, 22arxiv_flame, 23arxiv_diffusionunified} that apply DDPM to synthesize motions.
\cite{23arxiv_belfusion, 23arxiv_mdls} also explore latent diffusion models \cite{22cvpr_stablediffusion} for motion generation.
However, most works focus on text-conditioned \cite{22arxiv_motiondiffuse, 23arxiv_mofusion}, audio-driven \cite{23arxiv_ude} and music-driven \cite{23arxiv_edeg, 23arxiv_ude} motion generation while we target human-object interaction.
In this work, we apply transformer DDPM \cite{20nips_ddpm} to a multi-stage framework, which separately generates trajectories and synthesizes motions.

There are some concurrent works \cite{23arxiv_scenediffuser} that apply DDPM to synthesize motions in a scene.
However, \cite{23arxiv_scenediffuser} is designed for short-term motion generation (around $2$ seconds) while we target long-term human motion generation (longer than $10$ seconds). 
Different from existing work \cite{22arxiv_hmd, 23arxiv_scenediffuser}, we employ DDPM in a multi-stage framework, where trajectories and motions are separately predicted.
\section{Methods}
\label{sec:method}
Given the object $\mathbf{I}$ and the starting point $\mathbf{s}$, 
our goal is to synthesize 3D human motions $\{(\mathbf{r}_i, \boldsymbol{\theta}_i)\}^N_{i=1}$ with human-object interactions.
$\mathbf{r}_i$ is the root trajectory, and the $\boldsymbol{\theta}_i$ indicates local pose at $i$-th frame.

We design a hierarchical motion generation framework as shown in \figref{fig:overview}.
First, we employ GoalNet\cite{21iccv_samp} to predict an interaction goal on the object.
Then, we generate the goal pose (\secref{subsec:goalpose}) to explicitly model the human-object interaction. 
Next, our milestone generation module (\secref{subsec:milestone}) estimates the length of the milestones, produces the trajectory of the milestones from the starting point to the goal, and places milestone poses.
Therefore, the long-range motion generation is decomposed into combinations of short-range motion synthesis.
Finally, we design a motion generation module (\secref{subsec:motion}) to synthesize the trajectory between the milestones and infill the motions.
\subsection{Goal pose generation}
\label{subsec:goalpose}
We call the pose in which a person interacts with an object and remains stationary a goal pose.
To synthesize diverse human motion, we first generate a goal pose interacting with the object following \cite{22cvpr_goal, 22eccv_saga, 22cvpr_tdns}.
Most methods \cite{20cvpr_psi, 203dv_place, 21cvpr_posa} generate human poses using the cVAE model.
They project the poses to the standard normal distribution in the continuous space \cite{14iclr_vae}.
Empirically, we find that cVAE models do not perform well in our setting.
To overcome this challenge, we introduce VQ-VAE \cite{17nips_vqvae, 19nips_vqvae2} to model the data distribution which exploits a discrete representation to cluster the data in a finite set of points \cite{22eccv_posegpt}.
We hypothesize that limited data of goal pose from the SAMP dataset \cite{21iccv_samp} can always be clustered by VQ-VAE but may not be enough for learning a continuous latent space for VAE \cite{22cvpr_autosdf}.
In addition, based on the observation that different human poses may share similar properties \cite{21iccv_scatd, 22tog_motionpuzzle} (e.g., humans may sit with different hand positions but the same leg positions), we split joints into $L$ $(L=5)$ different non-overlapping groups like MotionPuzzle \cite{22tog_motionpuzzle}.
\begin{figure}[t]
  \centering
   \includegraphics[width=1.0\linewidth]{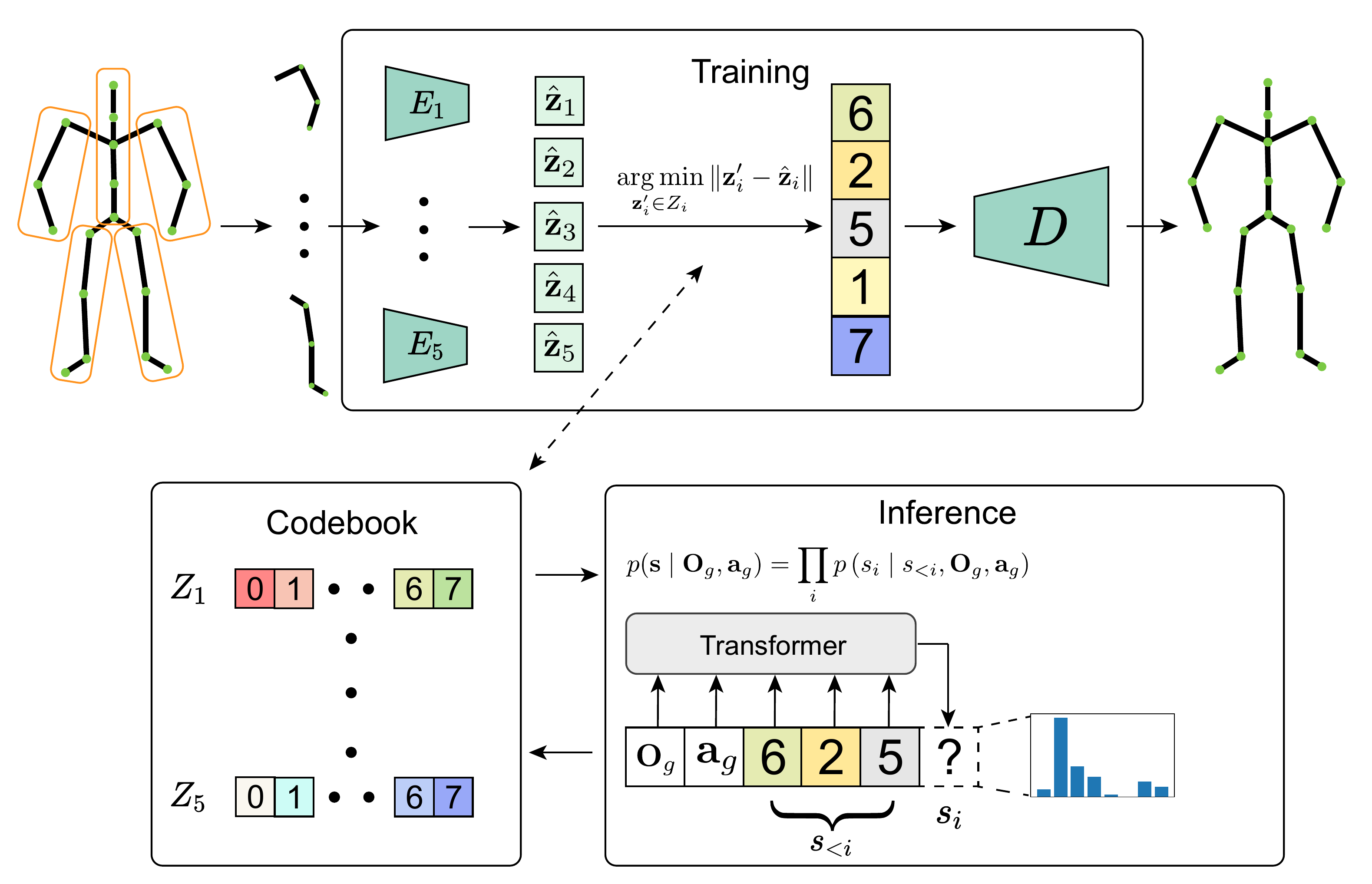}
   \caption{\textbf{Part VQ-VAE.} Part VQ-VAE first splits the skeleton into multiple parts and learns the codebooks separately. The composition of different parts is subsequently modeled with the autoregressive prediction model.}
   \label{fig:part_vqvae}
\end{figure}
\paragraph{Quantization.}
As shown in \figref{fig:part_vqvae}, the goal pose $\boldsymbol{\theta}_g$ is split into separate joint groups as $\boldsymbol{\theta}_g = \{\boldsymbol{\theta}_{gi}\}^L_{i=1}$.
Then a discrete codebook $\mathbf{Z}_i$ with a list of vectors compares the output $\hat{\mathbf{z}}_i$ from the encoder $E_i$ to find the closest vector in Euclidean distance.
The $L$ vectors with minimal distances will be concatenated and fed into a shared decoder $D$ to reconstruct $\boldsymbol{\theta}_{g}$. The loss function is defined as: 
\begin{equation}
    \resizebox{0.88\hsize}{!}{$
\begin{aligned}
    \mathcal{L}(\boldsymbol{\theta}_g, D(\mathbf{z}))=\|\boldsymbol{\theta}_g-D(\mathbf{z})\|_2^2+\sum_{i=1}^L\|s g[E_i(\boldsymbol{\theta}_{gi})]-\mathbf{z}_i\|_2^2 \\ +\beta\sum_{i=1}^L\|s g[\mathbf{z}_i]-E_i(\boldsymbol{\theta}_{gi})\|_2^2,
\end{aligned}
    $}
\end{equation}
where 
\begin{align}
    &\mathbf{z} = [\mathbf{z}_1, ..., \mathbf{z}_L], \\
    &\mathbf{z}_i=\underset{\mathbf{z}_i^{\prime} \in Z_i}{\arg \min }\left\|\mathbf{z}_i^{\prime}-\hat{\mathbf{z}}_i\right\|,\\
    &\hat{\mathbf{z}}_i = E(\boldsymbol{\theta}_{gi}).
\end{align}
Here, the term $sg[\cdot]$ donotes the stop-gradient operator, and $\|sg[\mathbf{z}_i]-E_i(\boldsymbol{\theta}_{gi})\|_2^2$ is the commitment loss \cite{19nips_vqvae2} controlled by a weighting factor $\beta$.
\paragraph{Generation.}
With $E_i$ and $D$ available, we can represent $\boldsymbol{\theta}_g = \{\boldsymbol{\theta}_{g1}, ..., \boldsymbol{\theta}_{gL}\}$ by a sequence of the part-based codebook indices. 
To be more specific, we use $E_i$ to extract the feature from $\boldsymbol{\theta}_{gi}$ and find the closest vector $\mathbf{z}_i \in \mathbf{Z}_i$. 
Then we use $s_i \in \{0, ..., |\mathbf{Z}_i| - 1\}$ to indicate the index of $\mathbf{z}_i$ in $\mathbf{Z}_i$. 
Therefore, $\boldsymbol{\theta}_g$ can be represented by $\mathbf{s} = \{s_1, ..., s_L\}$.

To generate a natural goal pose, we convert the problem to predict a sequence of indices that can represent $\boldsymbol{\theta}_g$.
We formulate the inference as a conditional auto-regressive process and employ a transformer \cite{17nips_attention} to learn to predict the distribution of possible indices \cite{21cvpr_taming}.
The condition contains the environment around the goal $\mathbf{O}_g$, and the action $\mathbf{a}_g$.
%
%
Following NSM \cite{19tog_nsm} and COUCH \cite{22eccv_couch}, a cylindrical volume of a pre-defined radius and height is created around the goal.
Within this volume, spheres are uniformly sampled and the occupancies corresponding to the object of these spheres are calculated.
Then, these occupancies are flattened
to form a feature vector denoted as $\mathbf{O}_g$.
$\mathbf{a}_g$ is a vector indicating the action type.
These variables are fed into the transformer as tokens. 
Our target is to learn the likelihood of the sequence:
\begin{equation}
    p(\mathbf{s} \mid \mathbf{O}_g, \mathbf{a}_g)=\prod_i p\left(s_i \mid s_{<i}, \mathbf{O}_g, \mathbf{a}_g\right).
\end{equation}
After predicting the indices, we map them back to their corresponding codebook entries to get the quantized features $\mathbf{z} = [\mathbf{z}_1, ..., \mathbf{z}_L]$, which are fed into the decoder $D$ to generate the goal pose $\boldsymbol{\theta}_g$.
\subsection{Milestone generation}
\label{subsec:milestone}
Based on the starting pose and the goal pose, we can generate the milestone trajectory and synthesize the local poses at the milestones.
Following \cite{22arxiv_motiondiffuse, 22arxiv_hmd}, we build a transformer DDPM \cite{20nips_ddpm} and apply it to generate the milestones for better quality. 
Because the length of motion data is unknown and can be arbitrary (e.g., the human could quickly walk towards the chair and sit down or sit after walking around the chair slowly), we predict the length of milestones, denoted by $N$.
Then we synthesize $N$ milestone points and place the local poses on these points.
\paragraph{Transformer DDPM.}
Here we first briefly introduce DDPM \cite{20nips_ddpm}, which is learned to reverse a diffusion process. 
Formally, the diffusion model \cite{15icml_diffusion} is defined as a latent variable model of the form $p_\theta\left(\mathbf{x}_0\right):=\int p_\theta\left(\mathbf{x}_{0: T}\right) d \mathbf{x}_{1: T}$, where $\mathbf{x}_0 \sim q\left(\mathbf{x}_0\right)$ is the data and $\mathbf{x}_1, \ldots, \mathbf{x}_T$ are the latents. 
$p_\theta\left(\mathbf{x}_{0: T}\right)$ is formulated as a Markov chain as:
\begin{equation}
    p_\theta\left(\mathbf{x}_{0: T}\right):=p\left(\mathbf{x}_T\right) \prod_{t=1}^T p_\theta\left(\mathbf{x}_{t-1} \mid \mathbf{x}_t\right),
\end{equation}
\begin{equation}
    p_\theta\left(\mathbf{x}_{t-1} \mid \mathbf{x}_t\right):=\mathcal{N}\left(\mathbf{x}_{t-1} ; \boldsymbol{\mu}_\theta\left(\mathbf{x}_t, t\right), \mathbf{\Sigma}_\theta\left(\mathbf{x}_t, t\right)\right).
\end{equation}
Diffusion models approximate posterior $q\left(\mathbf{x}_{1: T} \mid \mathbf{x}_0\right)$ as a Markov chain that gradually adds Gaussian noise to the data with variance schedules given by $\beta_t$:
\begin{equation}
    q\left(\mathbf{x}_{1: T} \mid \mathbf{x}_0\right):=\prod_{t=1}^T q\left(\mathbf{x}_t \mid \mathbf{x}_{t-1}\right),
\end{equation}
\begin{equation}
    q\left(\mathbf{x}_t \mid \mathbf{x}_{t-1}\right):=\mathcal{N}\left(\mathbf{x}_t ; \sqrt{1-\beta_t} \mathbf{x}_{t-1}, \beta_t \mathbf{I}\right).
\end{equation}
In contrast to adding noises on $\mathbf{x}_0$ sequentially, DDPM formulates the diffusion process as:
\begin{equation}
    q\left(\mathbf{x}_t \mid \mathbf{x}_0\right)=\mathcal{N}\left(\mathbf{x}_t ; \sqrt{\bar{\alpha}_t} \mathbf{x}_0,\left(1-\bar{\alpha}_t\right) \mathbf{I}\right),
\end{equation}
where $\alpha_t=1-\beta_t$ and $\bar{\alpha}_t=\prod_{s=1}^t \alpha_s$. 
Hence, we can generate $\mathbf{x}_t$ by sampling a noise $\epsilon$ as the training data. 
DDPM employs a neural network to model $p_\theta\left(\mathbf{x}_{t-1} \mid \mathbf{x}_t\right)$ and the inference is to gradually denoise $\mathbf{x}_t$ from $t=T$ to $t=1$ where $\mathbf{x}_T \sim \mathcal{N}(\mathbf{0}, \mathbf{I})$. 

\begin{figure}[t]
  \centering
   \includegraphics[width=1.0\linewidth]{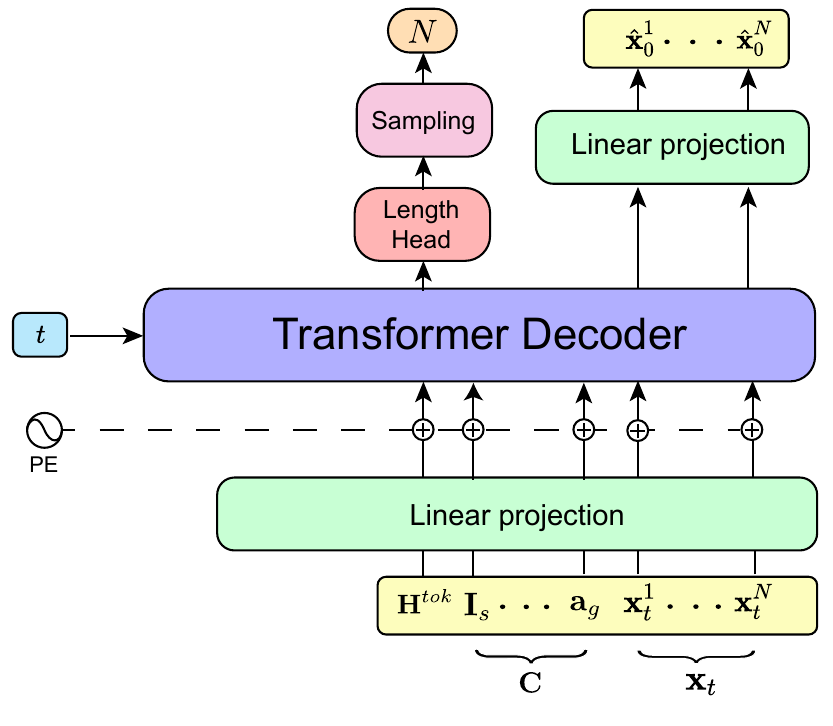}
   \caption{\textbf{Overview of transformer DDPM for milestone generation.} The model first takes the length token $\mathbf{H}^{tok}$ and the condition $\mathbf{C}$ as the input to predict the data length. Then it constructs the noise sequence $\mathbf{x}^{1:N}_T$ with length $N$. In the diffusion process, it is fed with $\mathbf{C}$ and the sequence $\mathbf{x}^{1:N}_t$ at time-step $t$ to predict the target $\hat{\mathbf{x}}^{1:N}_0$. For other sub-modules, we remove the length prediction head.}
   \label{fig:transformer_ddpm}
\end{figure}

Like existing works \cite{22arxiv_motiondiffuse, 22arxiv_flame} that apply DDPM in the motion domain, we employ a transformer decoder \cite{17nips_attention} as our architecture of DDPM.
The transformer takes the noise $\mathbf{x}_t$ and the condition $\mathbf{C}$ as input. The condition $\mathbf{C}$ means variables related to the generation, which will be described in detail in each subsection.
The diffusion time-step $t$ is in the sinusoidal position embeddings form \cite{17nips_attention} and is injected into each block in the transformer.
Different from existing works \cite{22arxiv_hmd, 22arxiv_motiondiffuse} that assume the length of the motion sequences is already given, we insert a parallel branch to estimate the length of milestones by taking the length token $\mathbf{H}^{tok}$ and the condition $\mathbf{C}$ as the input, as shown in \figref{fig:transformer_ddpm}.
The length prediction head is an MLP and predicts a multinomial distribution over discrete length indices $\{1, 2, ..., N_{max}\}$ like \cite{22cvpr_humanml3d}, where $N_{max}$ represents the length of the longest sequences for training.
We use cross-entropy loss as the loss function.
At inference time, we sample the length $N$ from the estimated distribution.
Next, we construct $N$ milestones as the input to the transformer.
We predict the denoised data $\hat{\mathbf{x}}_0$ based on the condition $\mathbf{C}$ by the DDPM $f$. 
This is formulated as $\hat{\mathbf{x}}_0 = f\left(\mathbf{x}_t, t, \mathbf{C} \right)$, and the training loss is defined as:
\begin{equation}
    \resizebox{0.88\hsize}{!}{$
        \mathcal{L}=\mathrm{E}_{t \in[1, T], \mathbf{x}_0 \sim q\left(\mathbf{x}_0\right), \epsilon \sim \mathcal{N}(\mathbf{0}, \mathbf{I})}\left[\|\mathbf{x}_0 - \hat{\mathbf{x}}_0\|\right].
    $}
\end{equation}

In our setting, we use a two-step strategy like \cite{21cvpr_slt, 22cvpr_tdns}, where two transformer DDPMs are applied to first generate the milestone points and then synthesize the local pose at each milestone.
\paragraph{Generation of milestone points.}
The milestone points are conditioned on the object, the information of the starting point, and the goal.
We then define the condition as:
\begin{equation}
    \mathbf{C_\mathbf{m}} = \{\mathbf{I}_s, \mathbf{I}_g, \mathbf{O}_s, \mathbf{O}_g, \mathbf{g}, \mathbf{s}\},
\end{equation}
where $\mathbf{I}_s$ and $\mathbf{I}_g$ are the object representation relative to the starting point and the goal. 
Similar to other methods \cite{19tog_nsm, 21iccv_samp, 22eccv_couch}, the object representation is modeled by an $8 \times 8 \times 8$ grid with their positions and occupancies.
Following NSM \cite{19tog_nsm} and COUCH \cite{22eccv_couch}, we also explicitly model the occupancies around the starting point and the goal with $\mathbf{O}_s$ and $\mathbf{O}_g$ (the same form as the occupancy feature described in \secref{subsec:goalpose}).
The information of goal $\mathbf{g}$ is defined as $\{\mathbf{r}_g, \mathbf{a}_g, \boldsymbol{\theta}_g\}$, where $\mathbf{r}_g$ means the goal position and orientation in the starting point coordinate system,
$\mathbf{a}_g$ is the target action label at the goal,
and $\boldsymbol{\theta}_g$ is the goal pose.
$\mathbf{s} = \{\mathbf{a}_s, \boldsymbol{\theta}_s\}$ represents the action labels $\mathbf{a}_s$ and the pose $\boldsymbol{\theta}_s$ at the starting point.

The target is to predict the milestone $\{\mathbf{m}_1, ..., \mathbf{m}_N\}$ with length $N$. 
Following NSM \cite{19tog_nsm}, a bi-directional scheme is employed for the milestone points generation where we predict the roots of milestones in both the starting point coordinate system and the goal coordinate system. 
The final predicted roots are blended from these two kinds of outputs. 
The milestone point $\mathbf{m}_i$ is defined as:
\begin{equation}
\label{equ:milestone}
\mathbf{m}_i = \{\mathbf{r}^{b}_{i}, \mathbf{c}_i, \mathbf{w}_i\}.
\end{equation}
The representation is similar to previous work \cite{21iccv_samp, 19tog_nsm, 22eccv_couch}, where $\mathbf{r}^{b}_{i}$ indicates the root position and forward directions relative to the starting point and the goal, $\mathbf{c}_{i}$ is a label vector indicating the contact between the environment and the body, and we use a high-dimensional feature vector $\mathbf{w}_i$ to encode the character state at the milestone following \cite{19tog_nsm, 21iccv_samp, 22eccv_couch}.
The details of these variables are provided in the supplementary material.
We use transformer DDPM $f_\mathbf{m}$ to predict the length and synthesize milestone points.
\paragraph{Generation of milestone poses.}
Previous works \cite{21cvpr_slt, 22cvpr_tdns} separately place the poses on the sparse points along the path and infill the motions between, which may lead to unnatural transitions between these poses.
On the contrary, we generate the local pose at each milestone with transformer DDPM to build the temporal dependency. 
With the help of the generated milestone points, we can access the accurate spatial relationship $\mathbf{I}_i$ between the object and the character and the ego-centric environment occupancies $\mathbf{O}_i$ \cite{22eccv_couch} at the $i$-th milestone. 
The milestone poses also depend on the milestone state including $\{\mathbf{c}_i, \mathbf{w}_i\}$ at the $i$-th milestone. 
Consequently, we define the condition $\mathbf{C_\mathbf{k}}$ as a combination with starting pose $\boldsymbol{\theta}_s$, goal pose $\boldsymbol{\theta}_g$, and frame-wise condition $\boldsymbol{\gamma}_i$ at the $i$-th milestone as: 
\begin{align}
    & \mathbf{C_\mathbf{k}} = \{\boldsymbol{\theta}_{s}, \boldsymbol{\theta}_{g}, \boldsymbol{\gamma}_1, ..., \boldsymbol{\gamma}_N\},\\
    &\text{where} \quad \boldsymbol{\gamma}_i = \{\mathbf{I}_i, \mathbf{O}_i, \mathbf{c}_i, \mathbf{w}_i\}.
    \label{equ:milestonepose}
\end{align}
The local poses at the milestones are predicted by the transformer DDPM $f_\mathbf{k}$ without the length prediction head.
\subsection{Motion generation}
\label{subsec:motion}
Instead of predicting motions frame-by-frame \cite{19tog_nsm, 21iccv_samp, 22eccv_couch}, our approach hierarchically synthesizes the full sequence based on the generated milestones.
We follow \cite{20eccv_lhmp, 21cvpr_slt, 22cvpr_tdns} to first generate the trajectory and then synthesize the motions.
Specifically, within two consecutive milestones, we first complete the trajectory.
Then, the motions are infilled with the guidance of successive milestone poses.
The two steps are accomplished using two transformer DDPMs (described in \secref{subsec:milestone}), respectively.
For each step, we carefully design the condition of DDPM to generate the target output.
\paragraph{Trajectory completion.} 
For the trajectory completion between the milestones $\mathbf{m}_i$ and $\mathbf{m}_{i+1}$, we assume that it is only conditioned on the milestones and the object.
Thus we define the condition as:
\begin{equation}
    \mathbf{C_\mathbf{r}} = \{\mathbf{I}_{i}, \mathbf{I}_{i+1}, \mathbf{O}_{i}, \mathbf{O}_{i+1}, \mathbf{m}_{i}, \mathbf{m}_{i+1}, \mathbf{t}^{i+1}_i\},
\end{equation}
where $\mathbf{I}_i$ indicates the object representation relative to the milestone $i$ and $\mathbf{O}_i$ denotes the ego-centric occupancies (the same form as the occupancy feature described in \secref{subsec:goalpose}) around the $i$-th milestone like NSM \cite{19tog_nsm} and COUCH \cite{22eccv_couch}. 
$\mathbf{m}_i$ has been shown in \equref{equ:milestone}. 
$\mathbf{t}^{i+1}_{i}$ represents the position and orientation of the $(i + 1)$-th milestone in the $i$-th milestone's coordinate system.
Between two consecutive milestones, we generate the trajectory with a length of 2 seconds, which is $61$ frames.
Similar hyperparameters can be found in previous methods \cite{21cvpr_slt, 22cvpr_tdns}.

The target output is the trajectory between two consecutive milestones.
Similar to the milestone point, we synthesize the trajectory in the $j$-th frame with a bi-directional scheme.
The trajectory is composed of a set of points that have the same representation as the milestone in \equref{equ:milestone}.
We generate the trajectory with a transformer DDPM $f_\mathbf{r}$ similar to the one in \secref{subsec:milestone}.
Since we assume the trajectory length between two milestones is fixed, the DDPM $f_\mathbf{r}$ does not have the length prediction head.
\paragraph{Motion infilling.}
To synthesize the long-range motion, we convert a long sequence into several fixed-length short sequences with the help of milestone points and milestone poses. 
For a sub-sequence between the consecutive milestone poses, our goal is to generate the missing local poses over the trajectory. 
The generated motion has to satisfy the trajectory and naturally transits from a milestone to the next milestone.
Like milestone pose generation, we use the same representation of frame-wise condition in \equref{equ:milestonepose}.
The condition is defined as:
\begin{equation}
    \mathbf{C_\mathbf{p}} = \{\boldsymbol{\theta}^1, \boldsymbol{\theta}^{61}, \boldsymbol{\gamma}^1, ..., \boldsymbol{\gamma}^{61}\}, 
\end{equation}
where $\boldsymbol{\theta}^1$ and $\boldsymbol{\theta}^{61}$ are the local poses of two consecutive milestones.
By taking these inputs, we generate smooth motions using another transformer DDPM $f_\mathbf{p}$ without the length prediction head.

\section{Experiments}
\label{sec:experiments}
\subsection{Implementation details}
\label{exp:implementation}
We train the part VQ-VAE and transformer DDPM models with the Adam optimizer \cite{14arxiv_adam}. 
All the models are trained with a fixed learning rate of $0.0001$ with batch size $256$. 
The remaining details are in the supplementary material.
\subsection{Datasets and evaluation metrics}
\label{exp:metric}
\paragraph{Test setting.}
Our experiments are conducted on the SAMP \cite{21iccv_samp}, COUCH \cite{22eccv_couch}, and NSM \cite{19tog_nsm} datasets.
Provided with a starting point, a starting pose, an object, and an endpoint, the virtual human is asked to approach the object, interact with it, and leave to reach the endpoint.
The ablation studies are conducted on the SAMP dataset.

\paragraph{Metrics.}
Following the previous method \cite{21iccv_samp}, we calculate the Fréchet distance (FD) between the generated and ground-truth motions to measure the motion quality.
We also conduct user studies and each sequence is evaluated by at least $3$ users with scores ranging from $1$ to $5$.
In addition, we calculate the penetration ratio \cite{21cvpr_slt, 203dv_place, 20cvpr_psi} and foot sliding \cite{18tog_mann, 20tog_motionvae} to show the physical plausibility between the 3D object and the synthesized motions. 
We compute the Average Pairwise Distance (APD) \cite{20eccv_dlow, 21cvpr_mojo} to evaluate the diversity.
Specifically, we calculate the APD of synthesized motion, the character's pose during object interactions, and trajectories.
Following previous work \cite{19tog_nsm, 21iccv_samp}, we calculate PE (positional errors) and RE (rotational errors) to indicate the precision of object interactions.
For each test object, we generate multiple sequences.
More details are included in the supplementary material.
\subsection{Comparison with other methods}
\label{exp:comparison}
\paragraph{Results on the SAMP dataset.} 
On the SAMP dataset \cite{21iccv_samp}, we compare our method with online methods SAMP \cite{21iccv_samp} and MoE \cite{18tog_mann}.
As our method is offline, we also implement and modify offline methods SLT \cite{21cvpr_slt} and TDNS \cite{22cvpr_tdns} to our setting.
For SLT \cite{21cvpr_slt}, we employ $\text{A}^*$ \cite{68_astar} to plan a path and select points along the path as subgoals to form the input for SLT.
More details about the implementation of \cite{21cvpr_slt, 22cvpr_tdns} are in the supplementary material.
Since MoE often fails to finish the action, we do not calculate the penetration ratio for it.
\begin{table*}[tb]
\begin{center}
\scriptsize{
\resizebox{0.7\linewidth}{!}{
\renewcommand{\arraystretch}{1.3}
    \setlength\tabcolsep{2.0pt}
        \begin{tabular}{l|ccccccccc}
        \hline
        Method & FD $\downarrow$ & User study $\uparrow$ & $\text{APD}_{M}\uparrow$ & $\text{APD}_{P}\uparrow$ & $\text{APD}_{T}\uparrow$ & PE$\downarrow$ & RE$\downarrow$ & Penetration$\downarrow$ & Sliding$\downarrow$ \\
        \hline
        MoE \cite{18tog_mann} & 74.33 & 2.76 & 3.50 & 2.63 & 52.46 & $\infty$ & $\infty$ & - & 1.68\\
        SAMP \cite{21iccv_samp} & 57.34 & 2.86 & 3.63 & 3.05 & 63.18 & 3.44 & 4.12 & 6.98 & 1.02\\
        $\text{SLT}^*$ \cite{21cvpr_slt} & 68.83 & 2.30 & 3.08 & 2.66 & 40.13 & 1.77 & 1.60 & 4.28 & 1.71\\
        $\text{TDNS}^*$ \cite{22cvpr_tdns} & 46.60 & 2.90 & 3.68 & 3.40 & 66.04 & 0.45 & 0.39 & 5.14 & 0.94\\
        Ours & \textbf{22.34} & \textbf{3.62} & \textbf{4.06} & \textbf{4.52} & \textbf{91.38} & \textbf{0.39} & \textbf{0.32} & \textbf{4.00} & \textbf{0.50}\\
      \hline
    \end{tabular}
    }
}
\end{center}
\caption{\textbf{Quantitative results on the SAMP dataset.} $\text{SLT}^*$ and $\text{TDNS}^*$ mean we modify and implement them on the SAMP dataset. The subscript ``$M$", ``$P$" and ``$T$" stand for ``Motion", ``Pose" and ``Trajectory". PE and RE represent the positional error and rotation error. Sliding denotes foot sliding. $\infty$ means the method failed to reach the goal.} 
\label{tab:samp}
\end{table*}
As shown in \tabref{tab:samp}, our approach outperforms other methods in terms of lower FD, higher user study scores, and higher APD. 
Furthermore, our method achieves much higher trajectory diversity than SAMP \cite{21iccv_samp}. 
Although TNDS \cite{22cvpr_tdns} proposes Neural Mapper (NM) which combines $\text{A}^*$ \cite{68_astar} and cVAE, the diversity of the generated trajectory is inferior to our method, as indicated by $\text{APD}_T$. 
\paragraph{Results on the COUCH dataset.} 
\begin{table}[tb]
\begin{center}
\scriptsize{
\resizebox{1.0\linewidth}{!}{
\renewcommand{\arraystretch}{1.3}
    \setlength\tabcolsep{2pt}
        \begin{tabular}{l|ccccccc}
        \hline
        Method & FD$\downarrow$ & User study $\uparrow$ & $\text{APD}_{M}\uparrow$ & $\text{APD}_{P}\uparrow$ & $\text{APD}_{T}\uparrow$ & Penetration$\downarrow$ & Sliding$\downarrow$ \\
        \hline
        NSM \cite{19tog_nsm} & 118.98 & 2.99 & 0 & 0 & 0 & 8.20 & 0.59\\
        SAMP \cite{21iccv_samp} & 160.12 & 1.94 & 0.89 & 0.18 & 4.69 & 4.94 & 0.72\\
        COUCH \cite{22eccv_couch} & 127.19 & 3.05 & 1.41 & 0.66 & 23.48 & 7.43 & \textbf{0.37}\\
        $\text{SLT}^*$ \cite{21cvpr_slt} & 93.46 & 3.01 & 1.68 & 1.17 & 2.86 & 3.90 & 1.85 \\
        $\text{TDNS}^*$ \cite{22cvpr_tdns} & 71.72 & 3.29 & 2.70 & 2.19 & 16.98 & 5.12 & 1.17 \\
        Ours & \textbf{56.35} & \textbf{4.27} & \textbf{3.22} & \textbf{2.30} & \textbf{64.96} & \textbf{3.54} & 0.55\\
      \hline
    \end{tabular}
    }
}
\end{center}
\caption{\textbf{Quantitative results on the COUCH dataset.} 
$\text{SLT}^*$ and $\text{TDNS}^*$ mean we modify and implement them.} 
\label{tab:couch}
\end{table}
As our target is to synthesize diverse motions instead of controlling the characters, we only evaluate the motion quality on the COUCH dataset \cite{22eccv_couch}. 
\tabref{tab:couch} shows that our method outperforms all baselines. 
Our approach achieves much higher $\text{APD}_T$ than other methods.
We observe that $\text{APD}_T$ of TDNS \cite{22cvpr_tdns} is higher than the approaches \cite{21iccv_samp, 21cvpr_slt} that employ deterministic $\text{A}^*$ \cite{68_astar}, but much lower than our method.
Although COUCH \cite{22eccv_couch} exhibits lower foot sliding than our method, it may sometimes get stuck, resulting in lower foot sliding since the character does not move.
\paragraph{Results on the NSM dataset.} 
\begin{table}[tb]
\begin{center}
\scriptsize{
\resizebox{1.0\linewidth}{!}{
\renewcommand{\arraystretch}{1.3}
    \setlength\tabcolsep{2.0pt}
        \begin{tabular}{l|ccccccccc}
        \hline
        Method & FD$\downarrow$ & User study$\uparrow$ & $\text{APD}_{M}\uparrow$ & $\text{APD}_{P}\uparrow$ & $\text{APD}_{T}\uparrow$ & PE$\downarrow$ & RE$\downarrow$ & Penetration$\downarrow$ & Sliding$\downarrow$ \\
        \hline
        NSM \cite{19tog_nsm} & 90.39 & 3.95 & 0 & 0 & 0 & 1.72 & 0.40 & 6.43 & \textbf{0.69}\\
        SAMP \cite{21iccv_samp} & 68.86 & 3.77 & 1.21 & 0.11 & 20.01 & 4.77 & 4.84 & 7.83 & 1.38\\
        Ours & \textbf{57.02} & \textbf{4.04} & \textbf{2.62} & \textbf{0.93} & \textbf{62.23} & \textbf{1.01} & \textbf{0.28} & \textbf{4.85} & 0.80\\
      \hline
    \end{tabular}
    }
}
\end{center}
\caption{\textbf{Quantitative results on the NSM dataset.}} 
\label{tab:nsm}
\end{table}
On the NSM dataset \cite{19tog_nsm}, we compare our approach with SAMP \cite{21iccv_samp} and NSM \cite{19tog_nsm}.
\tabref{tab:nsm} shows that our method outperforms baselines. 
Compared with the deterministic method NSM, our approach could generate stochastic motion and diverse trajectories.
\paragraph{Results in a cluttered scene.}
\begin{figure}[tb]
  \centering
   \includegraphics[width=1.0\linewidth]{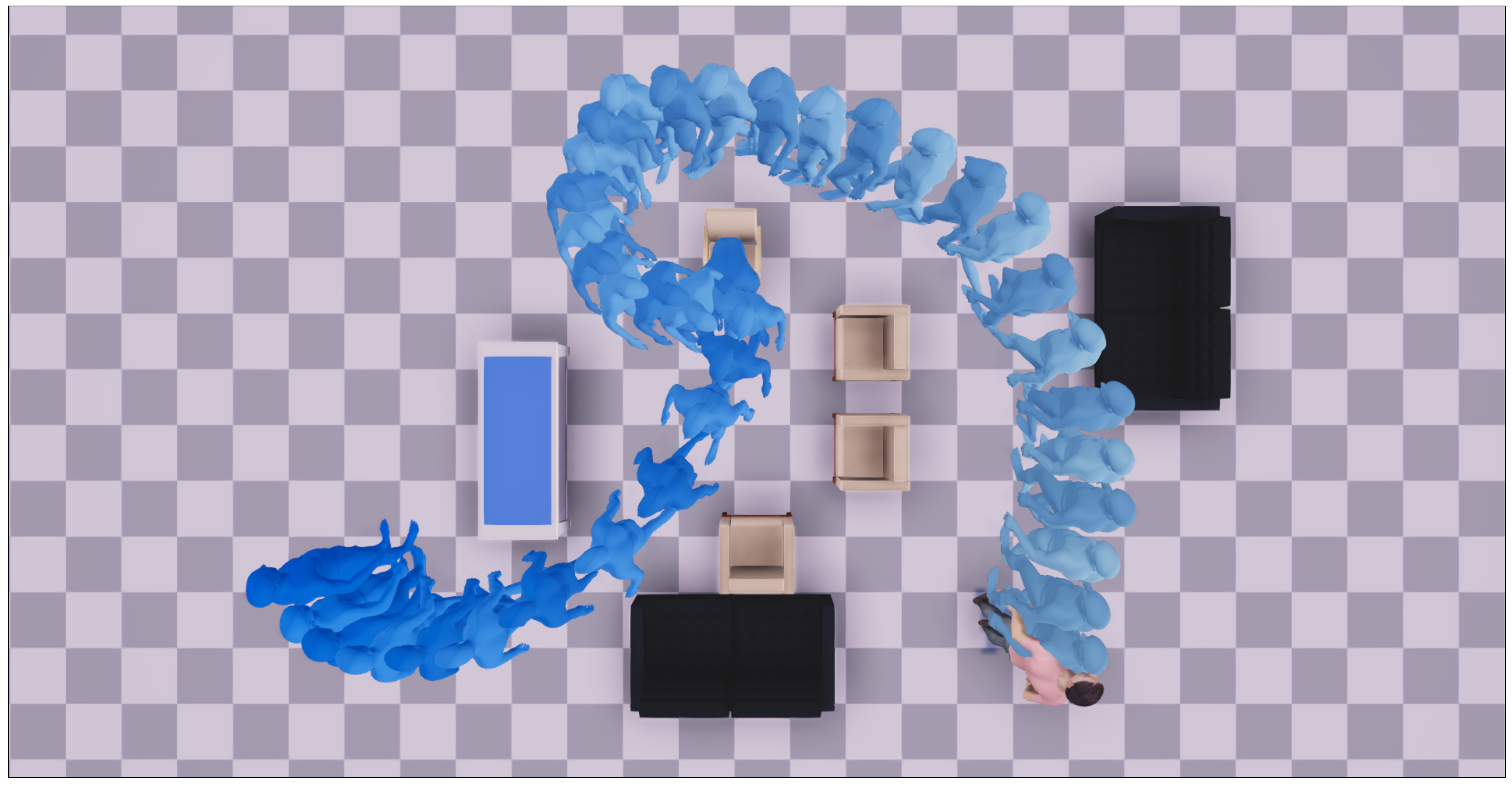}
   \caption{\textbf{Results in a cluttered scene.} Our method can generate motions that avoid obstacles in a cluttered scene.}
   \label{fig:scene}
\end{figure}    
We show our generated results in a cluttered scene in \figref{fig:scene}.
The percentage of frames with penetration is $3.8\%$ for our method and $4.9\%$ for SAMP.
More details are in the supplementary material.
\paragraph{Qualitative results.}
\begin{figure*}[tb]
  \centering
   \includegraphics[width=0.95\linewidth]{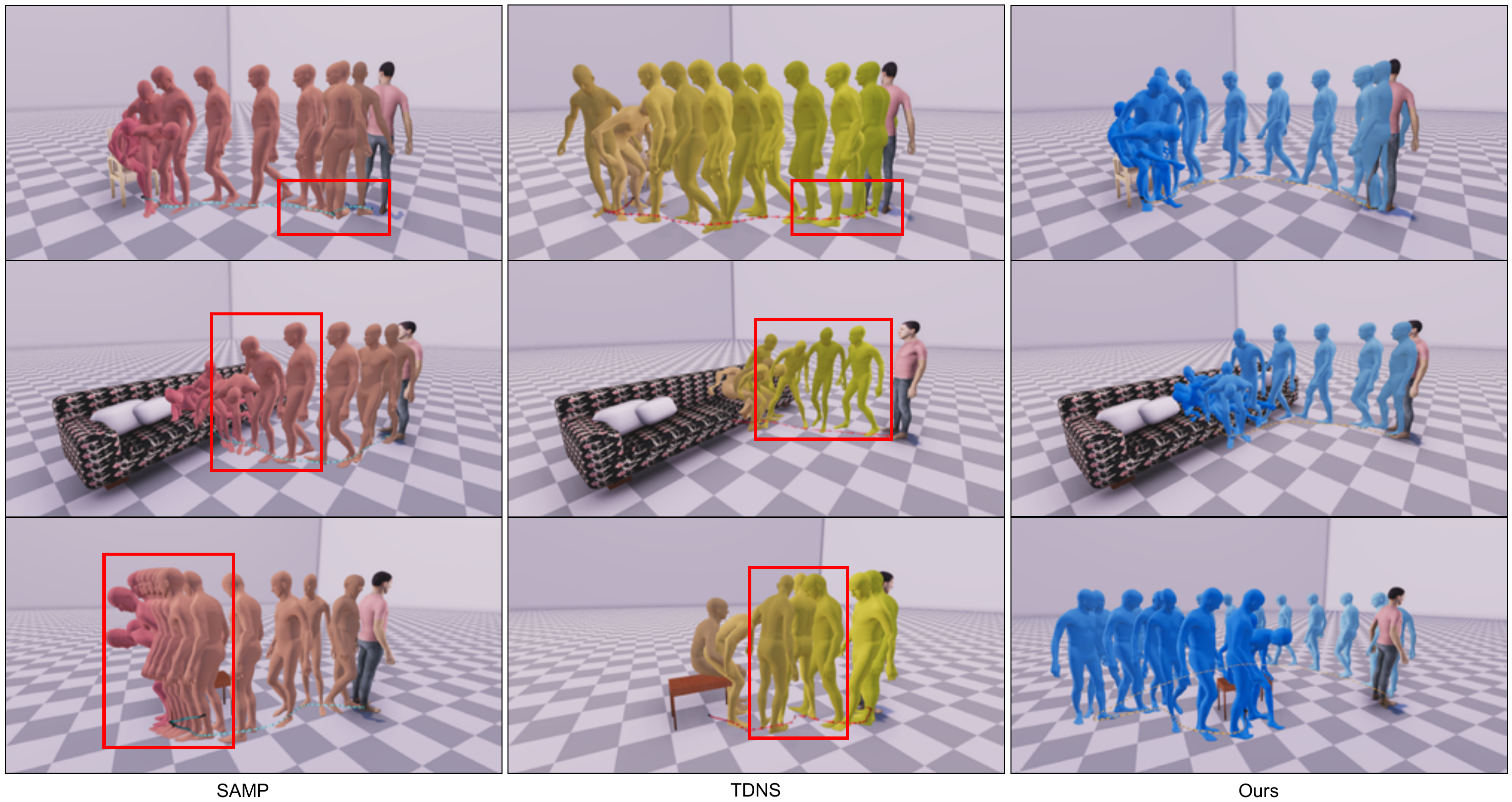}
   \caption{\textbf{Qualitative results on the SAMP dataset.} We compare our method with the baselines SAMP \cite{21iccv_samp} and TDNS \cite{22cvpr_tdns}. The failure cases are pointed out by the red rectangles. Specifically, the first row indicates that SAMP and TDNS tend to walk backward with more foot sliding. The second row shows that the baselines start to sit and lie down before reaching the object while our method synthesizes more natural results. The third row demonstrates that our framework has the capability to generate a long trajectory and walk naturally along the trajectory while SAMP gets stuck near the object and TDNS synthesizes unnatural motions. Lines on the floor indicate trajectories. The human with pink clothes indicates the start position. Darker color denotes later frames in the sequence.}
   \label{fig:visual}
\end{figure*}
\begin{figure*}[tb]
  \centering
   \includegraphics[width=0.95\linewidth]{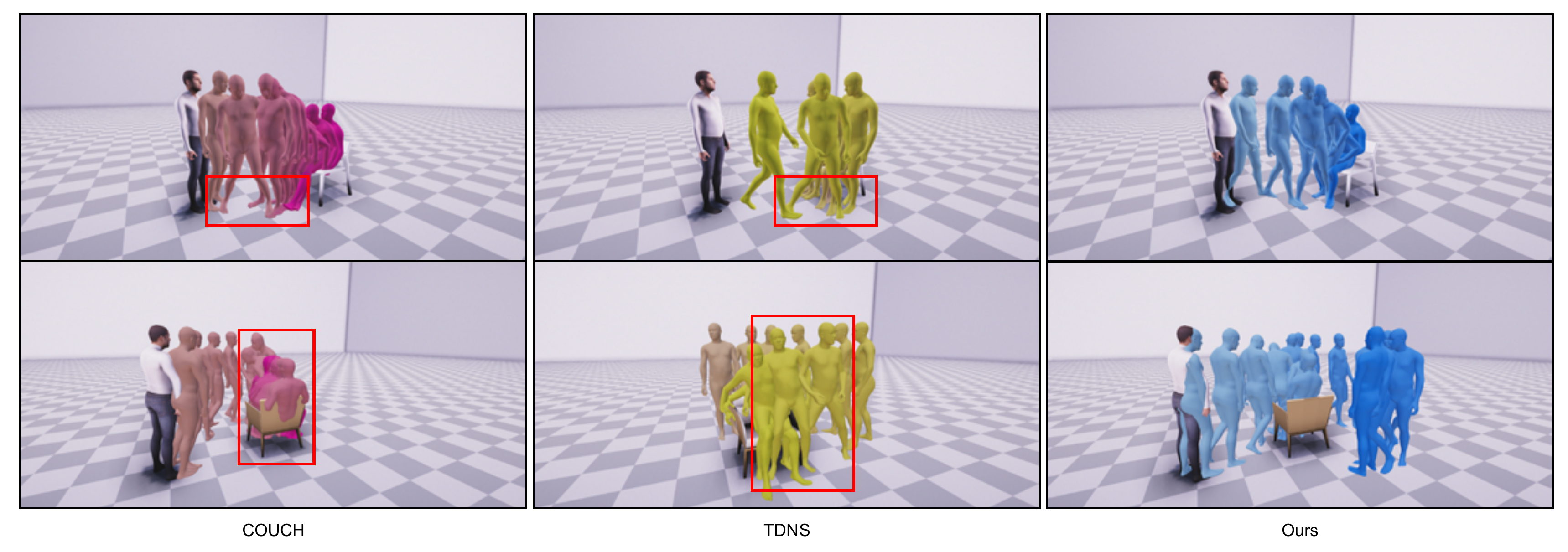}
   \caption{\textbf{Qualitative results on the COUCH dataset.} We compare our method with the baseline COUCH \cite{22eccv_couch} and TDNS \cite{22cvpr_tdns}. The failure cases are pointed out by the red rectangles. Specifically, the first row indicates that COUCH and TDNS tend to sit when the character is far from the object. The second row shows that COUCH may fail to stand up and TDNS may stand up unnaturally. To better visualize the results of TDNS in the second row, we only keep the frames in which the human stands up and goes to the endpoint.
   }
   \label{fig:visualcouch}
\end{figure*}
As demonstrated in \figref{fig:visual}, our approach achieves better results than baselines \cite{21iccv_samp, 22cvpr_tdns} on the SAMP dataset.
\figref{fig:visualcouch} compares our method with COUCH and TDNS on the COUCH dataset.
More qualitative results are in the supplementary material.

\subsection{Ablation study}
\label{exp:ablation}
\paragraph{Impact of each sub-module.} 
\begin{table}[tb]
\begin{center}
\scriptsize{
\resizebox{1.0\linewidth}{!}{
\renewcommand{\arraystretch}{1.3}
    \setlength\tabcolsep{2pt}
        \begin{tabular}{c|cccccc}
        \hline
        Variants & FD$\downarrow$ & $\text{APD}_{M}\uparrow$ & $\text{APD}_{P}\uparrow$ & $\text{APD}_{T}\uparrow$ & Penetration$\downarrow$ & Sliding$\downarrow$ \\
        \hline
         w/o GP & 31.07 & 3.21 & 2.46 & 85.21 & 4.10 & 0.49\\
        w/o MT &27.87 & 3.62 & 3.01 & \textbf{127.94} & 5.30 & 0.72\\
        w/o MP & 25.14 & 4.01 & 3.78 & 85.69 & 4.58 & \textbf{0.43}\\
        w/o TC & 36.77 & 2.76 & 2.42 & 83.26 & 4.34 & 0.63\\
        Ours & \textbf{22.34} & \textbf{4.06} & 4.52 & 91.38 & \textbf{4.00} & 0.50\\
      \hline
    \end{tabular}
    }
}
\end{center}
\caption{\textbf{Ablation study of the impact of sub-modules.} Although the variant without MT generates more diverse trajectories as shown in $\text{APD}_T$, the motion quality is much worse as indicated by the higher FD, penetration, and sliding. w/o: without. GP: goal pose generation. MT: milestone point generation. MP: milestone pose generation. TC: trajectory completion.} 
\label{tab:submodule}
\end{table}
To show the effectiveness of our hierarchical design, we evaluate our method against four variants where we remove one sub-module for each variant. 
\tabref{tab:submodule} indicates that each component improves performance. 
Generating a whole trajectory leads to more diverse trajectories but the motion quality is worse and has more foot sliding. 
This validates the necessity of the separate generation of trajectories and motions.
\paragraph{Goal pose generation.} 
We evaluate our goal pose generator with several variants, including cVAE, DDPM, and standard VQ-VAE.
For this evaluation, we only replace the goal pose generation module and keep the others the same.
\begin{table}[tb]
\begin{center}
\scriptsize{
\resizebox{1.0\linewidth}{!}{
\renewcommand{\arraystretch}{1.3}
    \setlength\tabcolsep{2.0pt}
        \begin{tabular}{c|cccccc}
        \hline
        Variants & $\text{FD}\downarrow$ & $\text{APD}_{M}\uparrow$ & $\text{APD}_{P}\uparrow$ & $\text{APD}_{T}\uparrow$ & Penetration$\downarrow$ & Sliding$\downarrow$ \\
        \hline
        cVAE & 27.06 & 3.36 & 3.27 & 90.52 & 4.36 & \textbf{0.47} \\
        DDPM & 34.22 & 3.75 & 3.37 & 84.76 & 4.31 & 0.49\\
        VQ-VAE & 24.77 & 3.78 & 3.87 & 89.19 & 4.27 & 0.48\\
        Part VQ-VAE & \textbf{22.34} & \textbf{4.06} & \textbf{4.52} & \textbf{91.38} & \textbf{4.00} & 0.50\\
      \hline
    \end{tabular}
    }
}
\end{center}
\caption{\textbf{Ablation study of goal pose generation.} We implement the goal pose module with different architectures.}
\label{tab:goalpose}
\end{table}
\begin{table}[tb]
\centering
\scriptsize{
\resizebox{1.0\linewidth}{!}{
    \setlength\tabcolsep{2pt}
    \renewcommand{\arraystretch}{1.3}
        \begin{tabular}{c|cccccc}
        \hline
        Variants & FD$\downarrow$ & $\text{APD}_{M}\uparrow$ & $\text{APD}_{P}\uparrow$ & $\text{APD}_{T}\uparrow$ & Penetration$\downarrow$ & Sliding$\downarrow$\\
        \hline
        Part VQ-VAE & 28.71 & 3.85 & 3.62 & 83.51 & 4.32 & 0.89 \\
        DDPM & \textbf{22.34} & \textbf{4.06} & \textbf{4.52} & \textbf{91.38} & \textbf{4.00} & \textbf{0.50}\\
      \hline
    \end{tabular}
    }
}
\caption{\textbf{Ablation study of part VQ-VAE for motion infilling.} We replace the DDPM as part VQ-VAE to predict motions.} 
\label{tab:partmotion}
\end{table}
\tabref{tab:goalpose} shows that part VQ-VAE generates more diverse poses than continuous latent space models. 
The comparison with standard VQ-VAE shows the necessity of our part design.
We also try part VQ-VAE for motion infilling, but the results in \tabref{tab:partmotion} show that its performance is worse.
\paragraph{Milestone generation.}
\begin{table}[tb]
\begin{center}
\scriptsize{
\resizebox{1.0\linewidth}{!}{
\renewcommand{\arraystretch}{1.3}
    \setlength\tabcolsep{2pt}
        \begin{tabular}{c|ccccccccc}
        \hline
        Variants & FD$\downarrow$ & $\text{APD}_{M}\uparrow$ & $\text{APD}_{P}\uparrow$ & $\text{APD}_{T}\uparrow$ & Penetration$\downarrow$ & Sliding$\downarrow$ \\
        \hline
        $\text{A}^*$ & 40.59 & 4.07 & 4.42 & 42.98 & 4.24 & 0.88\\
        NM \cite{22cvpr_tdns} & 40.54 & \textbf{4.15} & 4.31 & 52.74 & 4.15 & 0.96\\
        MT & \textbf{22.34} & 4.06 & \textbf{4.52} & \textbf{91.38} & \textbf{4.00} & \textbf{0.50}\\
      \hline
    \end{tabular}
    }
}
\end{center}
\caption{\textbf{Ablation study of milestone generation.} We compare our method with the variant based on the path generated by $\text{A}^*$ path planning \cite{68_astar}. 
NM: Neural Mapper \cite{22cvpr_tdns}. MT: milestone point generation.}
\label{tab:astar}
\end{table}
To further investigate the impact of milestones, we compare our approach with the variants that employ path-planning methods and select points along the path as milestones.
For this ablation, we implement $\text{A}^*$ and NM \cite{22cvpr_tdns} proposed by TDNS.
As demonstrated in \tabref{tab:astar}, the diversity of generated trajectories of $\text{A}^*$ \cite{68_astar} is much worse, and the motion quality drops significantly, as indicated by the lower $\text{APD}_T$ and higher FD.
The trajectory diversity of NM \cite{22cvpr_tdns} is better than $\text{A}^*$, but still worse than our method.
The reason why these variants perform poorly might be the low diversity of trajectory that affects the distribution of generated motions for calculating FD.
\paragraph{Motion infilling.}
\begin{table}[tb]
\centering
\scriptsize{
\resizebox{1.0\linewidth}{!}{
\renewcommand{\arraystretch}{1.3}
    \setlength\tabcolsep{2pt}
        \begin{tabular}{c|ccccccc}
        \hline
        Method & FD$\downarrow$ & $\text{APD}_{M}\uparrow$ & $\text{APD}_{P}\uparrow$ & Penetration$\downarrow$ & Sliding$\downarrow$ \\
        \hline
         ConvAE \cite{203dv_convae} & 26.50 & 3.95 & 4.33 & 4.82 & 0.89\\
         SLT \cite{21cvpr_slt} & 27.95 & 3.76 & 4.05 & 4.13 & 0.78\\
         Ours & \textbf{22.34} & \textbf{4.06} & \textbf{4.52} & \textbf{4.00} & \textbf{0.50}\\
      \hline
    \end{tabular}
    }
}
\caption{\textbf{Ablation study of motion infilling module.} We compare our method with other motion infilling methods.} 
\label{tab:infilling}
\end{table}
To validate our motion infilling module, we compare it with ConvAE \cite{203dv_convae} and SLT \cite{21cvpr_slt}.
We only replace the motion infilling module and keep the others the same.
The comparison of motion quality and diversity is shown in \tabref{tab:infilling} and our method outperforms ConvAE \cite{203dv_convae} and SLT \cite{21cvpr_slt} with lower FD.
\paragraph{DDPM vs. VAE.} 
To show the effectiveness of DDPM, we implement a cVAE variant, where we simply replace our transformer DDPM with transformer cVAE \cite{21iccv_actor, 22cvpr_tdns}.
\begin{table}[tb]
\begin{center}
\scriptsize{
\resizebox{1.0\linewidth}{!}{
    \setlength\tabcolsep{2pt}
    \renewcommand{\arraystretch}{1.3}
        \begin{tabular}{l|cccccc}
        \hline
        Arch & FD$\downarrow$ & $\text{R-APD}_{M}\uparrow$ & $\text{R-APD}_{P}\uparrow$ & $\text{R-APD}_{T}\uparrow$ & Penetration$\downarrow$ & Sliding$\downarrow$\\
        \hline
        cVAE & 29.32 & 3.65 & 3.58 & \textbf{109.55} & 5.06 & 1.12\\
        DDPM & \textbf{22.34} & \textbf{4.06} & \textbf{4.52} & 91.38 & \textbf{4.00} & \textbf{0.50}\\
      \hline
    \end{tabular}
    }
}
\end{center}
\caption{\textbf{Evaluation of the architecture for our generation framework.} Although cVAE based variant generates more diverse trajectories, the motion quality drops significantly as indicated by the much higher FD. Arch stands for the architecture type.}
\label{tab:ddpm}
\end{table}
As shown in \tabref{tab:ddpm}, although cVAE models could generate more diverse trajectories, their motion quality is far from satisfactory, indicated by the much higher value of FD.

~\\
\noindent
\textbf{Comparison with other diffusion models.}
\begin{table}[tb]
\centering
\scriptsize{
\resizebox{1.0\linewidth}{!}{
\renewcommand{\arraystretch}{1.3}
    \setlength\tabcolsep{2pt}
        \begin{tabular}{c|cccccc}
        \hline
        Variants & FD$\downarrow$ & $\text{APD}_{M}\uparrow$ & $\text{APD}_{P}\uparrow$ & $\text{APD}_{T}\uparrow$ & Penetration$\downarrow$ & Sliding$\downarrow$ \\
        \hline
        MDM \cite{22arxiv_hmd} & 38.69 & 4.01 & 3.77 & 72.67 & 5.19 & 0.70\\
        MDM \cite{22arxiv_hmd} + C & 23.97 & 4.02 & 4.13 & 89.80 & 4.87 & \textbf{0.39}\\
        FLAME \cite{22arxiv_flame} & 28.93 & 4.02 & 4.46 & 65.46 & 5.78 & 0.89 \\
        Ours & \textbf{22.34} & \textbf{4.06} & \textbf{4.52} & \textbf{91.38} & \textbf{4.00} & 0.50\\
      \hline
    \end{tabular}
    }
}
\caption{\textbf{Comparison with MDM and FLAME.} C denotes the frame-wise conditions.}
\label{tab:mdm}
\end{table}
Our approach stands out from architectures in MDM \cite{22arxiv_hmd} and FLAME \cite{22arxiv_flame} by incorporating frame-wise conditions.
\tabref{tab:mdm} demonstrates the significance of the frame-wise conditions.
\paragraph{More analyses and ablation studies.}
More detailed analyses and ablation studies of our design choices are provided in the supplementary material.
\subsection{Limitations}
\label{exp:limitation}
Although our method can generate diverse and natural motions, there are still some limitations.
Our method is offline and cannot be applied to interactive scenarios.
We assume that the objects are static and cannot handle moving objects.
The diffusion models require a long inference time.
It takes 7.13 seconds on average for a 720-frame sequence on a TITAN Xp GPU.
The slow speed might be solved by methods that could accelerate diffusion models \cite{21icml_improvedddpm, 22iclr_analyticdpm}.
\section{Conclusion}
\label{sec:conclusion}
In this work, we propose a novel hierarchical pipeline for motion synthesis of human-object interactions.
Our approach first generates the goal pose and then predicts a set of milestones.
Next, we synthesize motions with the guide of milestones.
Furthermore, we apply DDPM in our hierarchical pipeline.
We also show that our framework could generate more diverse and natural human-object interaction motions than other methods.

\section*{Acknowledgements}
\noindent
We thank Jintao Lu, Zhi Cen, Zizhang Li, and Kechun Xu for the valuable discussions.
This work was partially supported by the Key Research Project of Zhejiang Lab (No. K2022PG1BB01), NSFC (No. 62172364), and Information Technology Center and State Key Lab of CAD\&CG, Zhejiang University.

{\small
\bibliographystyle{ieee_fullname}
\bibliography{11_references}
}

\clearpage

\end{document}